%% file: main.tex
\documentclass[runningheads]{llncs}
\usepackage{eccv}

\usepackage{eccvabbrv}
\usepackage{graphicx}
\usepackage{booktabs}
\usepackage[accsupp]{axessibility}  
\usepackage{makecell}
\usepackage{twemojis}

\usepackage[table]{xcolor}
\definecolor{rex_blue}{RGB}{0,0,255}

\definecolor{claude_red}{RGB}{255,0, 0}

\definecolor{hint}{RGB}{150, 0, 255}

\newcommand{\ours}{Match-Any-Events}
\newcommand{\ourdata}{Event Cross Matching}
\newcommand{\ourdatashort}{ECM}
\newcommand{\ourdatasynth}{E-MegaDepth}

\definecolor{rankone}{RGB}{255, 210, 210}   
\definecolor{ranktwo}{RGB}{255, 225, 190}   
\definecolor{rankthree}{RGB}{255, 245, 200} 
\newcommand{\first}[1]{\cellcolor{rankone}\textbf{#1}}
\newcommand{\second}[1]{\cellcolor{ranktwo}#1}
\newcommand{\third}[1]{\cellcolor{rankthree}#1}

\usepackage[breaklinks,colorlinks,citecolor=eccvblue]{hyperref}
\usepackage{hyperref}
\usepackage{multirow}
\usepackage{orcidlink}

\begin{document}

\title{Match-Any-Events: Zero-Shot Motion-Robust Feature Matching \texorpdfstring{\\}{ } Across Wide Baselines for Event Cameras} 

\titlerunning{Match-Any-Events}

\author{Ruijun Zhang\inst{1}\orcidlink{0000-0001-5136-3205} \and
Hang Su\inst{2}\orcidlink{0000-0003-3365-4361} \and
Kostas Daniilidis\inst{3,4}\orcidlink{0000-0003-0498-0758} \and 
Ziyun Wang\inst{1}\orcidlink{0000-0002-9803-7949}}

\authorrunning{Zhang et al.}

\institute{Johns Hopkins University, Baltimore MD 21218, USA
\and
ShanghaiTech University, Shanghai, China
\and
University of Pennsylvania, Philadelphia PA 19104, USA
\and
Archimedes, Athena RC, Greece
}

\input{sec/teaser}
\input{sec/0_abstract}   
\input{sec/1_intro}
\input{sec/2_related}

\input{sec/3_method}
\input{sec/4_data_gen}
\input{sec/5_experiment}
\input{sec/6_conclusion}
\clearpage

%
%
\bibliographystyle{splncs04}
\bibliography{cleaned}
\end{document}

%% file: sec/teaser.tex
\maketitle
\vspace{-0.2cm}
\begin{center}
    \captionsetup{type=figure} 
    \includegraphics[trim=3.4cm 6.1cm 2.9cm 7.3cm, clip, width=\linewidth]{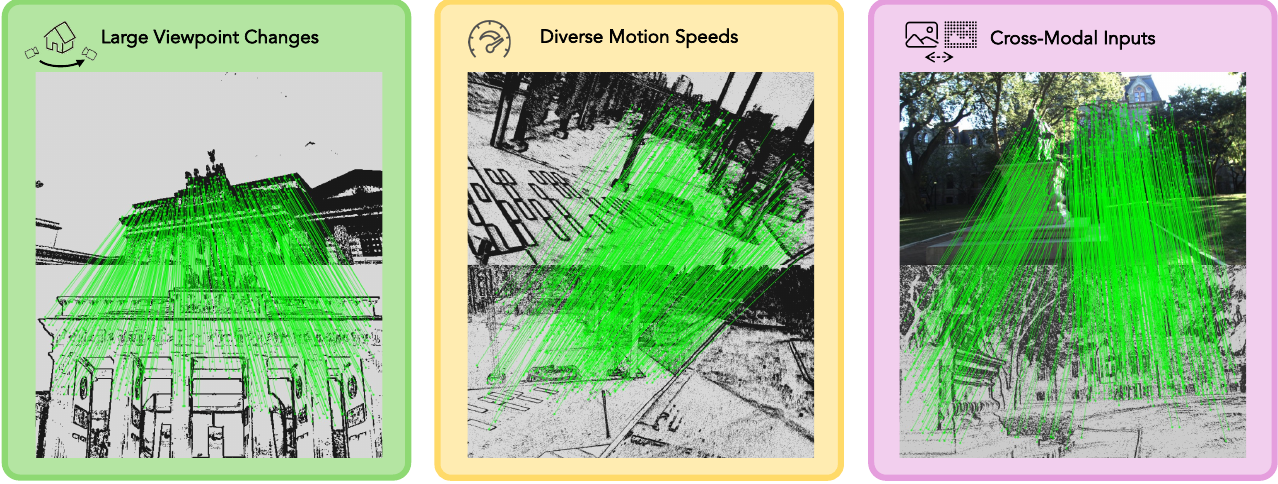}
    \caption{\ours{} can perform robust feature matching between two \textbf{arbitrary views} with different \textbf{motion profiles} and \textbf{across modalities}. 
    No fine-tuning is needed for new datasets.}
    \label{fig:teaser}
\end{center}
\vspace{-0.5cm}


%% file: sec/0_abstract.tex
\begin{abstract}
Event cameras have recently shown promising capabilities in instantaneous motion estimation due to their robustness to low light and fast motions. However, computing wide-baseline correspondence between two arbitrary views remains a significant challenge, since event appearance changes substantially with motion, and learning-based approaches are constrained by both scalability and limited wide-baseline supervision. 
We therefore introduce the first event matching model that achieves cross-dataset wide-baseline correspondence in a \textbf{zero-shot manner}: a single model trained once is deployed on unseen datasets without any target-domain fine-tuning or adaptation. 
To enable this capability, we introduce a motion-robust and computationally efficient attention backbone that learns multi-timescale features from event streams, augmented with sparsity-aware event token selection, making large-scale training on diverse wide-baseline supervision computationally feasible.
To provide the supervision needed for wide-baseline generalization, we develop a robust event motion synthesis framework to generate large-scale event-matching datasets with augmented viewpoints, modalities, and motions. Extensive experiments across multiple benchmarks show that our framework achieves a $\mathbf{37.7\%}$ improvement over the previous best event feature matching methods. \textit{Code and data are available at: \url{https://github.com/spikelab-jhu/Match-Any-Events}}.
\end{abstract}

%% file: sec/1_intro.tex
\section{Introduction}
\label{sec:intro}
Biological vision systems exhibit remarkable abilities to establish correspondence across different spatial and temporal scales. At short spatiotemporal intervals, instantaneous motions of objects are captured by optical flow processed within motion-sensitive areas, such as Middle Temporal (MT) area in the brain~\cite{born2005structure, gibson1950perception}. On the other hand, humans can easily establish correspondence between two distinct views with minimal overlaps, even in the absence of temporal continuity between two observations~\cite{dicarlo2012does, epstein1998cortical, murray2004attention, logothetis1996visual}. This intrinsic understanding of correspondence enables self-localization, loop-closure in odometry, and recognition of structural and semantic similarities between objects. Recently, frame-based computer vision algorithms have made substantial progress for both instantaneous and wide-baseline matching, driven by large-scale datasets and scalable network architectures~\cite{sarlin2020superglue, lindenberger2023lightglue, pautrat2023gluestick, sun2021loftr, wang2024efficient, wang2022matchformer, li2025edm}. \\

In contrast, biologically inspired event-based cameras have demonstrated \textbf{unbalanced} performance between \textbf{instantaneous} and \textbf{wide-baseline} correspondence. Event algorithms excel at estimating instantaneous motions due to their high temporal resolution and dynamic range, allowing them to handle extremely dynamic and event deformable objects~\cite{zhu2017event, zhu2018ev, gehrig2020eklt, wang2022ev, wang2023event, wang2025continuous, wang2025beyond}. However, event-based methods have not shown generalizable performance in wide-baseline matching for several crucial factors. First, event data processing usually relies on a \textbf{hand-crafted} encoding function that depends on hyperparameters such as time interval and number of events. These parameters change the appearance of events based on motions, which prevents networks trained on certain motion speeds from generalizing to different motions. 
Second, we do not have sufficient scale for the current event-based matching networks and datasets. On the data side, existing event datasets provide narrow-baseline optical flow ground truth, annotated with SLAM-derived poses and point maps. In our experiments, networks trained with only such data cannot generalize to large viewpoint changes. The need for diverse motion augmentations further amplifies the data problem. On the model side, dense spatiotemporal attentions in event transformers increase the attention cost by $\mathcal{O}(T^2)$, where $T$ is event volume time resolution, making it computationally prohibitive to train networks on bigger datasets. Due to these issues, we have yet to develop a generalizable event-based matching network with good zero-shot performance on unseen data. In this work, we seek to fill this missing link in event-based matching from two different directions: \textbf{architecture design} and \textbf{supervision}. \\

We design an event transformer that explicitly separates spatial and temporal aggregation to make multi-timescale tokens tractable at scale. Particularly, events present unique challenges that break the assumptions underlying 2D transformers for matching. First, events form an inherently 3D \textbf{spatiotemporal representation} rather than a fixed 2D grid. Second, motion profiles vary drastically across pixels, from slow drift to rapid edges, producing \textbf{non-uniform sampling} in time. To this end, we introduce a motion-aware and separable feature encoder layer, which computes time and space feature aggregation efficiently. While naive multi-scale motion-aware layers would substantially increase the inference cost of transformers due to the large number of tokens, our proposed architecture mitigates this by decoupling spatial and temporal aggregation in a computationally efficient manner. Concretely, we use separable space and time attention to reduce the attention cost from $\mathcal{O}((THW)^2)$ to $\mathcal{O}(T(HW)^2 + HW\,T^2)$, which is the key enabling factor for training on millions of wide-baseline pairs rather than the narrow-baseline regimes used previously.
In addition, we introduce an adaptive sparsity-aware event token selection module (SETS) that adaptively prunes redundant tokens for matching. By allocating computation only to informative spatiotemporal regions and time steps, SETS further reduces the effective cost of multi-timescale processing while preserving matching accuracy.\\

On the supervision side, we contribute two new datasets that cover a much wider range of view changes for event-based matching: a synthetic \textbf{\ourdatasynth{}} dataset and a real \textbf{\ourdata{} (\ourdatashort{})} dataset. \ourdatasynth{} contains simulated events generated from the large MegaDepth~\cite{li2018megadepth} dataset, featuring extensive, wide-baseline view changes and motion profiles. For real-world evaluation, we carefully designed a hardware-synchronized hetero-stereo setup that captures images and events simultaneously. Our \ourdatashort{} dataset enables real-world evaluation by providing synchronized images and event streams with dense and precise correspondence annotations across wide baselines. Using SOTA multi-view geometric foundation models, we generate point maps of higher density and precision compared to classical structure-from-motion solutions. \\

Finally, we trained our model using the combined large-scale dataset, achieving state-of-the-art results in semi-dense matching and camera pose estimation for both in-domain data and unseen test in-the-wild data without any fine-tuning. Our main contributions are as follows:
\begin{itemize}
\item \textbf{Zero-shot wide-baseline event matching}. 
We formalize the setting where viewpoint and motion profiles vary jointly, and propose a single matcher that generalizes to unseen datasets without test-time tuning, supporting both event-to-event and event-to-image correspondence. 
\item \textbf{Efficient multi-timescale event aggregation}.
We introduce a motion-robust attention backbone that learns multi-timescale features via separable spatial–temporal aggregation, and further propose sparsity-aware event token selection to adaptively prune redundant tokens, resulting in efficient spatiotemporal transformers for events. 

\item \textbf{Wide-baseline supervision for events}.
We release two complementary datasets, \textbf{E-MegaDepth} (large-scale synthetic wide-baseline pairs with diverse motions) and \textbf{ECM} (real hetero-stereo event–image data with dense correspondences), enabling scalable training and rigorous evaluation. 
\end{itemize}

%% file: sec/2_related.tex
\section{Related Work}
\label{sec:formatting}

\textbf{Image-based Matching.}
Traditional image matching follows a detect-describe-match pipeline, where salient keypoints are extracted and encoded by handcrafted descriptors~\cite{lowe2004distinctive, bay2006surf, rublee2011orb}. 
Their invariance abilities made them the backbone of early SfM~\cite{schoenberger2016sfm} and SLAM systems~\cite{campos2021orb}, but they suffer from performance degradation in low-texture and fast motion scenes. 
Learning-based methods improved this pipeline through data-driven descriptors and joint detection-description, as shown in LIFT~\cite{yi2016lift}. SuperPoint~\cite{detone2018superpoint} popularized self-supervised keypoint learning, followed by ~\cite{revaud2019r2d2,tyszkiewicz2020disk,dusmanu2019d2}, which enhanced feature confidence and distinctiveness. Later correspondence estimation models like SuperGlue~\cite{sarlin2020superglue}, LightGlue~\cite{lindenberger2023lightglue}, and Gluestick~\cite{pautrat2023gluestick} leverage attention or graph reasoning to enforce geometric consistency, yet these detector-dependent approaches remain sparse and struggle in textureless or repetitive regions.
\\

Detector-free methods establish correspondences directly from dense feature maps without keypoint detection. LoFTR~\cite{sun2021loftr} pioneered a coarse-to-fine transformer for semi-dense matching, showing strong robustness in low-texture areas. Later variants such as ~\cite{wang2024efficient,wang2022matchformer,li2025edm} improved efficiency and multi-scale consistency through local correlation and lightweight architectures.
Beyond semi-dense matching, dense methods such as DKM~\cite{edstedt2023dkm}, COTR~\cite{jiang2021cotr}, and RoMA~\cite{edstedt2024roma} achieve pixel-level correspondence via global correlation and transformer reasoning across large viewpoints. These methods bridge image matching with optical flow and dense reconstruction, offering fine-grained details but having high computational cost. More recently, foundation models such as VGGT~\cite{wang2025vggt} have been developed to provide a unified approach to 3D prediction tasks. Our approach follows a semi-dense design that balances dense matches with high efficiency and competitive performance. \\

\textbf{Event-based Matching.}
Event-based methods follow a similar trajectory as their image counterparts. Early studies adapted classical descriptors such as corners~\cite{alzugaray2018asynchronous,Mueggler17BMVC,zhu2017event,ikura2024rate}, keypoints~\cite{alzugaray:BMVC20,lagorce2016hots,hu2022ecdt}, and edges~\cite{brandli2016elised, ikura2025lattice} to events. However, these handcrafted features remain highly noise-sensitive and lack repeatable correspondences over wide baselines. To better exploit spatio-temporal information, subsequent works encode events into grid-like or volumetric structures that preserve temporal details, including time surfaces~\cite{benosman2013event, zhu2018ev}, event volumes~\cite{gehrig2019end, zhu2019unsupervised, qu2024evrepsl, zhu2018ev, zhu2021eventgan, hamann2024motion, wang2022evac3d}, and feature fields~\cite{rodriguez2024s, das2025fast}.
Such representations laid the groundwork for downstream learning-based detection and matching models.
Building on these, recent methods have directly learned local features for event-based correspondence. EventPoint~\cite{huang2023eventpoint} learns both keypoint detection and dense descriptors through spatio-temporal consistency, achieving self-supervised training without human annotations. SD2Event~\cite{gao2024sd2event} further improved this framework by introducing a Contextual Feature Descriptor module to model long-range dependencies and Dynamic Keypoint Detector Learning module to adaptively detect keypoint. Zhu~\etal~\cite{zhu2025spatio} integrates multi-level temporal pyramids to enhance descriptor consistency and scale robustness in high-speed scenarios. In a more recent work, SuperEvent~\cite{burkhardt2025superevent} leverages image-based detectors on events to achieve robust detection and description under complex noise, and demonstrates strong performance in an event-based SLAM system. \\

\textbf{Cross-Modal Matching.}
With the growing adoption of hybrid stereo setups that combine event and RGB cameras, event-to-image matching has become crucial for multimodal fusion and 3D perception~\cite{wang2021stereo}. 
Hybrid systems benefit from the complementary properties of both sensors--dense spatial information from frames and high temporal resolution from events.
EKLT~\cite{gehrig2020eklt} explored event-RGB alignment by applying the classical KLT tracker on both events and frames. Beyond hybrid setups, event-based stereo matching itself has seen significant progress with improved disparity estimation~\cite{liu2022learning,piatkowska2017improved, zhu2018realtime} and spatial alignment in visual odometry~\cite{niu2025esvo2}. More recent works~\cite{kim2022real, cho2022selection, wang2024towards} focused on direct event-image correspondence by jointly optimizing cross-modal features between events and frames. Later works~\cite{lou2024zero, zhang2022data} extend these directions to zero-shot or wide-baseline configurations. At a broader scale, cross-modal matching frameworks, such as MINIMA~\cite{ren2025minima} and MatchAnything~\cite{he2025matchanything},  aim to learn foundational correspondence for diverse vision modalities. 
\\

%% file: sec/3_method.tex
\section{Preliminaries}

\textbf{Event Camera Data.}
Event cameras asynchronously capture per-pixel changes in logarithmic brightness. When the change in a pixel exceeds a contrast threshold $C$, the sensor outputs an event  $e_i = (x_i, y_i, t_i, p_i)$, where $(x_i, y_i)$ denotes the pixel coordinate, $t_i$ is the timestamp and $p_i\in \{+1, -1\}$ indicates the polarity of the brightness change. The resulting event stream $E = \{e_i\}^N_{i=1}$ encodes fine-grained motion and texture of the scene that are useful for robot navigation.

\paragraph{Problem Formulation.} 
We formally define the semi-dense event matching problem. Let $E^{\mathcal{A}}$ and $E^{\mathcal{B}}$ denote two event streams from viewpoints $\mathcal{A}$ and $\mathcal{B}$, respectively, which may differ by arbitrary baselines and motion profiles. Our task is to estimate a semi-dense correspondence field over this domain:
\begin{equation}
    \mathcal{M}, \mathcal{P} = f_{\theta}(E^{\mathcal{A}}, E^{\mathcal{B}}) \,,
\end{equation}
where $ f_{\theta}$ is a learnable matching function parameterized by network weights $\theta$, $\mathcal{P} \in [0, 1]^{H\times W}$ is the matching score map, and $\mathcal{M}(x, y) = (x^{'}, y^{'})$ is the dense correspondence map. Each valid location in view $\mathcal{A}$ where $\mathcal{P}(x,y)>\omega$ is mapped to its corresponding coordinate $(x^{'}, y^{'})$ in view $\mathcal{B}$.

\section{Method}

\begin{figure*}[ht]
    \centering
    \includegraphics[trim=3.8cm 4.9cm 3.3cm 5.6cm,clip,width=1.0\linewidth]{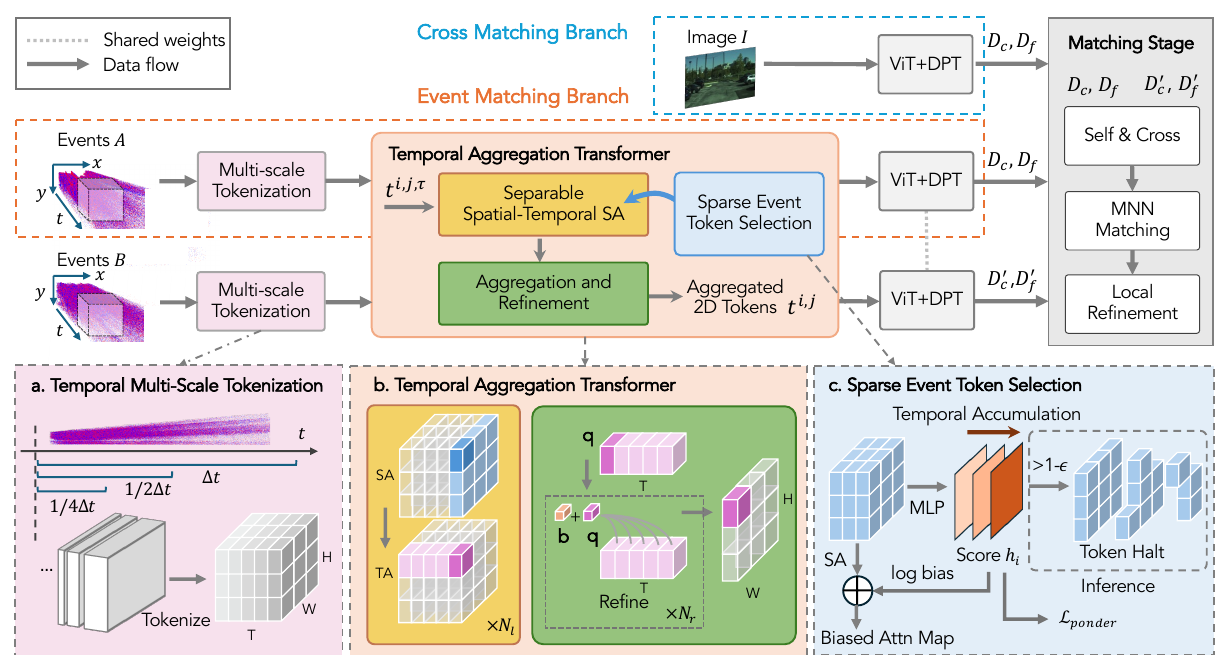}
    \caption{\textbf{Pipeline.} Event slices are binned to multi-scale voxel preserving rich temporal information (\cref{sec:representation}). The voxel inputs are tokenized and processed by a separable Temporal Aggregation Transformer module (\cref{sec:tau}), and then with a sparsity-aware token selection (\cref{sec:sets}) module. Multi-scale and dense features from ViT followed by DPT are used for iterative matching in the final MNN matching module (\cref{sec:matching}).
    }
    \label{fig:pipeline}
    \vspace{-.5cm}
\end{figure*}

Our method integrates event encoding, temporal aggregation, and dense matching into a unified, motion-robust end-to-end framework capable of predicting dense feature matches for both event-to-event and event-to-image matching tasks.
\subsection{Multi-timescale Event Representation}

\label{sec:representation}
To use scalable architectures for event data, such as vision transformers, the first step is to design compatible input representations that preserve the rich temporal and spatial information. Early works convert event streams into 2D image-like tensors, such as event frames and time surfaces~\cite{benosman2013event,gehrig2019end}. However, rich temporal information is lost when events are squashed into a single surface~\cite{zhu2018ev}. Subsequently, event voxel representations~\cite{zhu2018ev, baldwin2022time} emerged as a balanced solution between temporal resolution and computational efficiency. \\

We construct an event voxel logarithmically windowed in time~\cite{burkhardt2025superevent,hamann2025etap}, as illustrated in \cref{fig:pipeline}. Specifically, an event stream is divided into $B$ temporal bins, each representing a different accumulation time. Given $N$ events $\{ (x_i, y_i, t_i, p_i) \}_{i=1}^N$, the voxel volume is defined as
\begin{equation}
V(c, x, y) = \sum_i p_i \, k_b(x - x_i) \, k_b(y - y_i) \ \mathbb{I}(c > c_i^{*}) \,,
\label{eq:volume}
\end{equation}
where $c_i^{*} = \max(0, B-1+\log_2(\frac{t_i - t_1}{t_N - t_1})) $
and $k_b(a) = \max(0,\, 1 - |a|)$ denote a bilinear interpolation kernel, $\mathbb{I}(\cdot)$ is the indicator function, and $c_i^{*}$ is the scaled event timestamp. After voxelization, the representation is tokenized into a set of tokens $\mathbf{t} \in \mathbb{R}^{T\times H\times W\times D}$ that retain temporal information. However, processing these multi-timescale temporal tokens effectively for dense correspondence requires a specialized architecture.
To this end, we introduce a novel transformer that adaptively selects features across scales for high-quality matching.

\begin{table*}[t]
\centering

\caption{Quantitative comparison for event-to-event and event-to-image matching on ECM and M3ED.\colorbox{rankone}{Red}indicates the top performance, while\colorbox{ranktwo}{orange}and\colorbox{rankthree}{yellow}denote the second and third, respectively.}
\label{tab:main}
\renewcommand{\arraystretch}{1.2}
\setlength{\tabcolsep}{2pt}
\scriptsize
\begin{tabular}{c c c c c | c c c c }
\toprule
& \multicolumn{4}{c}{\makecell{\textbf{ECM Dataset}}} &
\multicolumn{4}{c}{\textbf{M3ED Dataset}~\cite{chaney2023m3ed}}  \\
& AUC@5° & AUC@10° & AUC@20° & Prec(\%) & AUC@5° & AUC@10° & AUC@20° & Prec(\%) \\
\rowcolor{gray!10}
& \multicolumn{8}{c}{\textbf{Event-to-Event Matching}} \\
M-A (Events) & 20.69 & 36.27 & 51.54 & \third{45.26} & 24.89  & 38.35  &  50.64 & \second{47.59} \\
M-A (E2V) & \second{41.68} & \second{59.71} & \second{73.01} & \second{50.05} & \second{38.04}  & \second{50.69}  &  \second{60.47} & \third{46.79} \\
VGGT (Events) & 1.86 & 6.74 & 17.45 & 12.06 & 15.90  & 24.24  & 33.70 & 20.20 \\
VGGT (E2V) & \third{31.44} & \third{52.03} & \third{68.40} & 43.41 & \third{28.99}  & \third{42.25}  & \third{52.46} & 36.29  \\
SuperEvent & 11.40 & 22.12 & 34.24 & 33.93 &  18.07 & 29.34  &  40.65& 39.43 \\
Ours & \first{54.61} & \first{72.24} & \first{82.67} & \first{68.90} & \first{52.99}  & \first{66.69}  &  \first{76.40} & \first{67.40} \\
\addlinespace[3pt]
\rowcolor{gray!10}
& \multicolumn{8}{ c}{\textbf{Event-to-Image Matching}} \\
\addlinespace[3pt]
M-A (Events) & \third{23.60}  & \third{38.93}  & \third{52.99}  &  \third{41.53} &  \third{26.35} & \third{38.92}  & \third{50.22} & \second{43.01}  \\
M-A (E2V) & \second{41.70}  & \second{59.60}  &  \second{72.09}  &  \second{49.68} & \second{31.92}  & \second{45.88}  & \second{56.92} & \third{39.07}  \\
VGGT (Events) &  2.18 & 6.32  & 13.17  & 8.29   &  4.26 & 8.47  &  14.23 & 9.34 \\
VGGT (E2V) &  18.10 & 34.25  & 50.75  & 30.94  & 13.17  & 21.70  & 29.85 & 21.02  \\
Ours & \first{48.58}  & \first{66.27}  & \first{78.30}  & \first{60.73}  & \first{54.97}  & \first{68.03}  &  \first{76.92} & \first{62.33} \\
\bottomrule
\end{tabular}
\end{table*}

\begin{table}[ht]
\centering

\caption{\label{tab:eds} \textbf{Pose estimation on EDS~\cite{hidalgo2022event}. } AUC is calculated with rotation errors.}
\begin{tabular}{lcccc}
\toprule
\textbf{Models} &\textbf{AUC@5°}& \textbf{AUC@10°} & \textbf{AUC@20°} \\
\midrule
RATE${}^\dag$~\cite{ikura2024rate} & 2.1 & 5.1 & 10.3 \\
EventPoint${}^\dag$~\cite{huang2023eventpoint} & 1.6 & 2.8 & 5.2  \\
Superevent~\cite{burkhardt2025superevent} & 25.4 & 37.5 & 49.0 \\
Ours & \textbf{40.4} & \textbf{56.2}& \textbf{68.8} \\
\bottomrule
\end{tabular}
\vspace{-.5cm}
\end{table}

\subsection{Temporal Aggregation Transformer}
\label{sec:tau}

The duration of an event sequence is often treated as a hyperparameter in event-based neural network design, affecting both training and inference performance. Short intervals preserve sharp spatial details but result in sparse and noisy representations, whereas longer intervals produce denser outputs at the cost of motion blur. While prior work~\cite{nam2022stereo} has proposed an auxiliary convolutional network, Concentrate Net, to recover a visually consistent 2D tensor, the temporal information is discarded due to averaging over time. To resolve this fundamental tradeoff, we introduce \textbf{Temporal Aggregation Transformer (TAg)}, which performs temporal attention and feature fusion within an end-to-end framework and directly learns from the training objective.
Our proposed module has two stages: Separable Spatial-temporal Attention and Temporal Aggregation Refinement. \\

\textbf{Separable Spatial-Temporal Attention.} Inspired by recent developments in video vision transformers ~\cite{arnab2021vivit, bertasius2021space, gritsenko2024end}, we first introduce a separable attention module that alternates between spatial and temporal dimensions of the input tokens $\mathbf{t} \in \mathbb{R}^{T\times H\times W\times D}$. Specifically, the spatial self-attention operates on tokens $\mathbf{t}^{\tau} \in \mathbb{R}^{ H\times W\times D}$ within each temporal bin $\tau$,  while the temporal self-attention attends to tokens $\mathbf{t}_{n}\in \mathbb{R}^{T\times D}$ across the temporal dimension, where $n$ is the spatial index. Intuitively, this encourages each embedding to gather information from its spatio-temporal neighbors, making the subsequent aggregation more robust. This design enables channel-wise and temporal feature fusion while reducing the computational cost from $O((T H W)^2)$ to $O(T (H W)^2 + H W T^2)$. We interleave spatial and temporal attention for $N_l$ times for fine feature extraction.
\\

\textbf{Temporal Aggregation and Refinement.} To address event-specific problems like
motion variance features under different binning window
sizes, we introduce an event-based component to integrate multi-temporal-scale event maps into a canonical
motion-invariant representation. Specifically, the second stage consolidates the information into a single feature map by querying the tokens in the first bin $\mathbf{t}^{\tau=0} \in \mathbb{R}^{H\times W\times D}$, which has the finest temporal resolution. The keys and values come from the remaining temporal bins within the same spatial indices. 
We directly query features from the bin with the finest temporal resolution, which contains sharp spatial details. 

A learnable bias $\mathbf{b}\in\mathbb{R}^D$ is added to the query to enhance feature quality when the finest bin contains too little information. Due to the attention-based aggregation, the model remains robust to the input event interval, as shown in our ablation experiments. 
To better understand TAg, we visualize the \textbf{attention weights} of the first aggregation attention layer in~\cref{fig:attention}. The model adaptively attends to fine-scale temporal bins in regions with fast motions, while shifting focus to coarse-scale temporal bins in regions with slow motions.
\begin{figure}[t]
    \centering
    \includegraphics[trim=1.7cm 7.3cm 1.1cm 1.9cm,clip,width=1\linewidth]{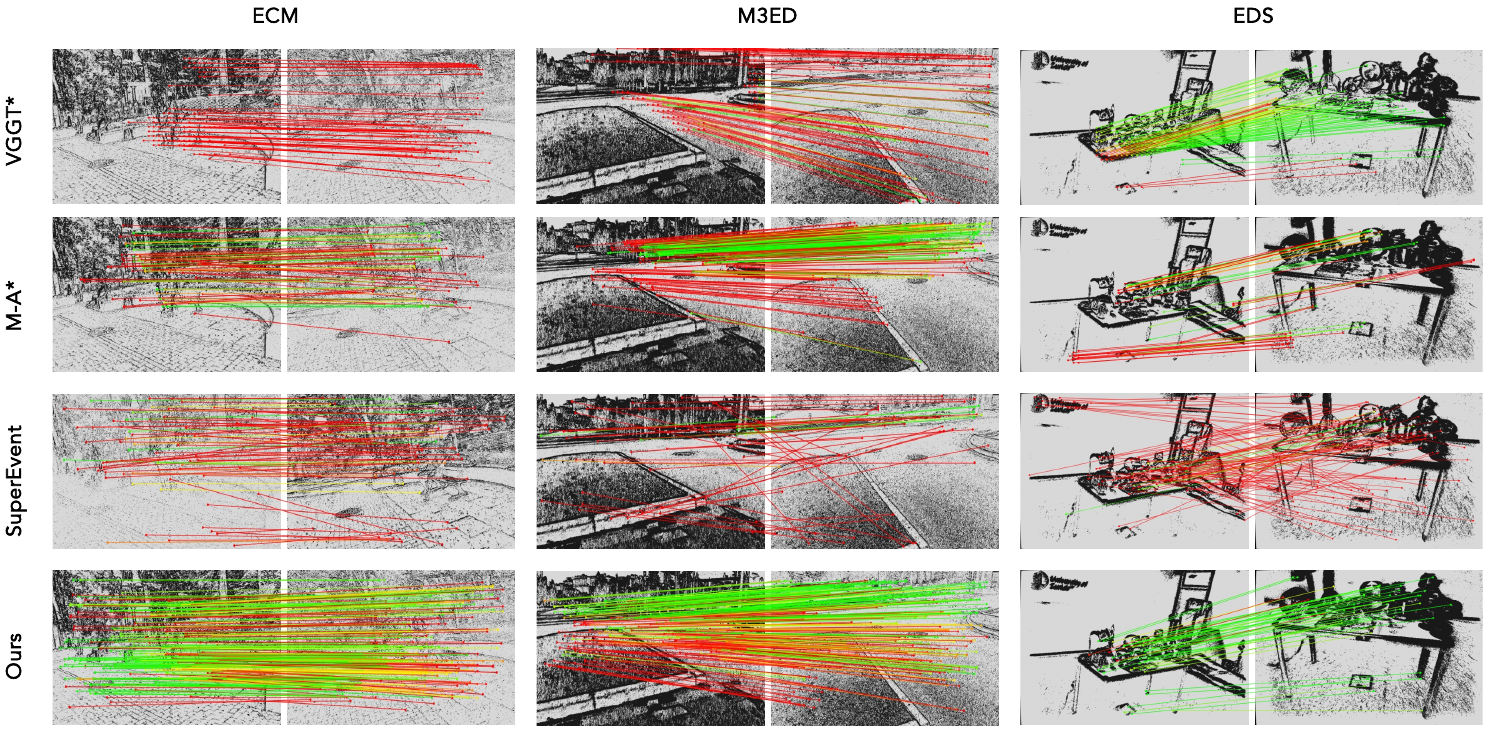}
    \caption{\textbf{Qualitative results of Match-Any-Events.} We compared with VGGT~\cite{wang2025vggt}, MatchAnything~\cite{he2025matchanything} and SuperEvent~\cite{burkhardt2025superevent}. \ours{} computes denser matches and achieves lower pose estimation errors.
    Green and Red indicate inliers and outliers for pose estimation respectively. Methods marked with $\star$ use event frames as input.}
    \vspace{-.5cm}
\end{figure}
\subsection{Sparsity-aware Event Token Selection (SETS)\label{sec:sets}}
\begin{figure*}[ht]
    \vspace{-.2cm}
    \centering

    \includegraphics[trim=0.9cm 11.7cm 1.2cm 4.3cm,clip,width=\linewidth]{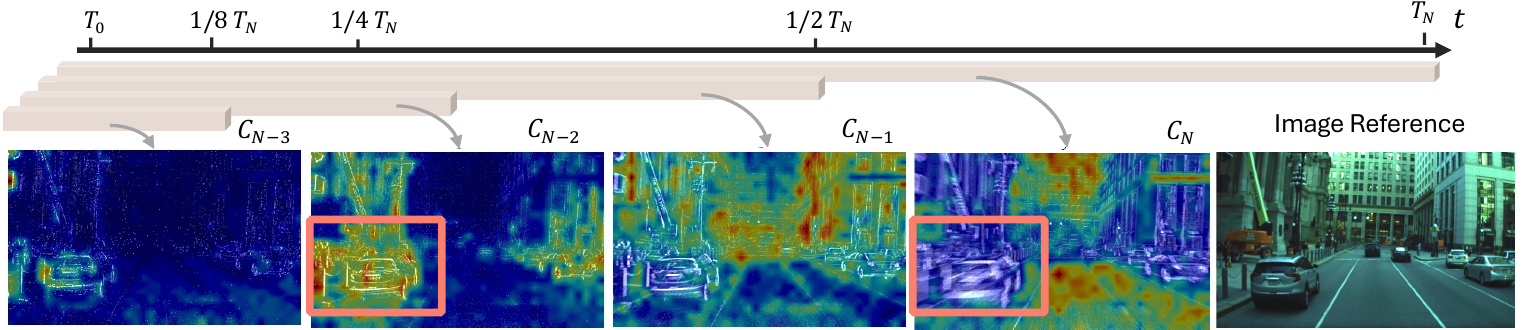}
    \caption{\textbf{Visualize Temporal Attention Weights.} 
    We visualize the attention weights of different time scales for the Aggregation and Refinement module. \textbf{Red} indicates high weight, while \textbf{blue} indicates low weight. The network attends to texture-rich regions at short temporal scales, and focuses on low-texture areas at longer temporal scales.
    }
    \label{fig:attention}
\end{figure*}

Although the separable attention design in~\cref{fig:attention} significantly reduces the computational cost, our further analysis shows that the dense spatial attention introduces most of the FLOPs. Specifically, computing spatial attention at every temporal resolution yields a computational complexity of $O(T(HW)^2)$. Inspired by ACT~\cite{graves2016adaptive}, we propose \textbf{Sparsity-aware Event Token Selection (SETS)}, an adaptive token selection module to prune uninformative event tokens due to the inherent sparse nature of events. \\

Assuming the average spatial sparsity of event data is $\alpha \in (0, 1]$, the computational complexity can be optimized to $O(T\alpha^2(HW)^2)$. Unlike previous token pruning techniques that rely on fixed, sub-optimal reduction ratios~\cite{rao2021dynamicvit,alvar2025divprune}, SETS learns the sparsity structure of the data without needing task-specific prior knowledge. For each token $t_n^\tau\in \mathbb{R}^{T\times(HW)\times C}$ with spatial-temporal information, we predict a halting score $h_n^\tau$ at each temporal step $\tau$ through an MLP followed by a sigmoid.
Scores are accumulated over temporal steps until they reach a threshold of $1-\epsilon$ at which point token processing halts. The variable $N_n$ tracks the number of steps taken prior to halting.
To ensure the mechanism remains completely differentiable and that the halting probabilities sum exactly to 1, we replace the final halting score with a remainder term $R_n = 1-\sum_{i=1}^{N_n-1}h_n^i$. \\

We then define the ponder loss $L_{\text{ponder}}$, combining the remainder term $R_n$ and a smoothing constant $N_n$, which encourages early halting and penalizes redundant temporal processing: $L_{ponder} = \frac{1}{HW}\sum_{n=1}^{HW}(N_n+R_n)$.
Rather than formulating the neural module as a mean-field model and averaging the outputs across layers in prior work~\cite{graves2016adaptive,yin2022vit}, we directly inject the accumulated halting score as a bias term into the spatial attention weights in each temporal resolution:

\begin{equation}
    \text{bias}_n^\tau = \begin{cases} 
    \log \left(1-\sum_{i=1}^{\tau-1}h_n^i\right), & \tau\leq N_n \\ 
    -\infty, & N_n<\tau\leq T 
    \end{cases}
    \label{eq:bias_injection}
\end{equation}

We compute a bias map from the bias term $\text{bias}_n^\tau$ and inject it into the raw spatial attention map. By introducing this bias, tokens with high halting scores are actively suppressed during the attention operation. 
Consequently, the combination of the matching loss and ponder loss acts as a push-pull mechanism, effectively balancing predictive performance with computational efficiency. 

\begin{figure*}[t]
    \centering
    \includegraphics[trim=8.1cm 13.5cm 4.5cm 5.0cm,clip,width=\linewidth]{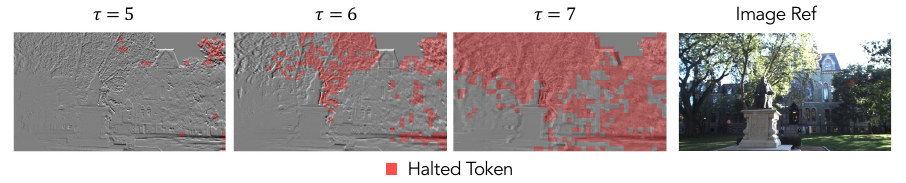}
    \caption{\textbf{Visualizing Sparsity-Aware Token Selection.} We highlight the halted tokens at each temporal step using red patches. As the temporal information accumulates, an increasing number of tokens in spatially blurry regions are halted.
    }
    \label{fig:sparsity}
    \vspace{-.5cm}
\end{figure*}
\subsection{Matching Stage}
\label{sec:matching}

 We address the semi-dense matching problem by progressively refining correspondences in a coarse-to-fine manner. Cross and self attentions are applied between two coarse feature maps (Stride$=$14) in an alternating fashion. The coarse features are correlated to build a score matrix $S$, on which dual-softmax is applied to obtain a matching probability $P$. The matching correspondence indices $(x_i,y_j)$ are selected using Mutual Nearest Neighbors (MNN):
 \begin{equation}
 \label{eq:mnn}
 (x_i, y_j) = (\arg\max_x S(x, y_j),\arg\max_y S(x_i, y))
 \end{equation}
 The indices from coarse matching are used to crop fine features into local patches, where fine correspondences are determined with MNN similar to coarse features. Finally, a $3\times 3$ window is cropped, and the predictions are refined by computing the expected values over the local region. \\
 
 \textbf{Loss functions.} The coarse-level matching loss is formulated by minimizing the cross-entropy loss over the ground-truth matching matrix $M_{gt}$:
\begin{align}
    L_c = -\frac{1}{|M^{gt}_c|}\sum M^{gt}_c\log P_c.
\end{align}
Similarly, the fine-level loss is also supervised on the correlated matching matrix in fine scale:
$L_f = -\frac{1}{|M^{gt}_f|}\sum M^{gt}_f\log P_f.$
We use a refinement stage to generate sub-pixel level prediction. The refinement prediction is supervised by $L_l$, which is defined as an $\mathcal{L}_{2}$ loss between the predicted coordinate and ground-truth coordinate. The final loss consists of the losses for the coarse-level and the fine-level: $L = L_c + \alpha L_f + \beta L_l + \gamma L_{ponder}$.

%% file: sec/4_data_gen.tex
\section{\ourdatasynth{} and \ourdatashort{} Datasets ~\label{sec:data}}

\textbf{\ourdatasynth{}} builds on MegaDepth~\cite{li2018megadepth}, a large-scale internet photo collections dataset with poses and dense depth annotations. In contrast to the predominantly driving oriented content of existing event datasets, MegaDepth offers rich diversity in viewpoint changes, making it suitable for evaluating wide-baseline matching. We construct a new synthetic event-based dataset, \ourdatasynth{}, by warping the images using a randomly sampled 6-DoF transformation. By interpolating these transformations, we synthesize high frame-rate video, which is further converted into event streams using Vid2E~\cite{gehrig2020video}. During event generation, we randomly select contrast sensitivity $C \sim \mathcal{U}(0.16,0.34)$ following~\cite{klenk2024deep}. We generate approximately 3 million pairs of event streams for training. \\

\textbf{Event Cross Matching Dataset (ECM)}. Unlike previous real event outdoor datasets collected under driving motions~\cite{orchard2015converting,zhu1801multi,mueggler2017event}, which lack motion diversity and viewpoint variations that are important for generalizable matching. To address this gap, we collected a new dataset using a calibrated and synchronized multi-camera system including a Prophesee Gen4 camera (1280$\times$720) and a Flir RGB camera (1920$\times$1080). For annotation, we provide COLMAP~\cite{schoenberger2016sfm} poses and bundle-adjusted VGGT~\cite{wang2025vggt} depth. Each scene is recorded with two different trajectories with diverse motion patterns and large viewpoint changes. 

%% file: sec/5_experiment.tex
\section{Experiments}

In this section, we describe the experimental details, including datasets, baselines and metrics~\cref{sec:baseline}, and report evaluation results thoroughly. We focus on two tasks: event-to-event matching and event-to-image matching.

\subsection{Data, Baselines, Metrics and Implementation}
\label{sec:baseline}
\textbf{Datasets.}
We evaluate our method on M3ED~\cite{chaney2023m3ed}, EDS~\cite{hidalgo2022event}, and \ourdatashort{} quantitatively. Please refer to~\cref{sec:data} for details of \ourdatashort{}. The M3ED dataset provides multi-sensor recordings from ground, aerial, and legged robots equipped with event cameras, RGB cameras, LiDAR, and IMU. It provides synchronized event streams. To obtain ground-truth correspondences, we project 3D point clouds between frames using known sensor extrinsics and poses. We sample 20,000 pairs for training and 1,000 pairs for testing. EDS records indoor scenes under varying illumination using a beam splitter that captures synchronized events and images. Note that our model is not trained on either \ourdatashort{} or EDS, which enables real zero-shot evaluation.
\\

\textbf{Protocol.} On \textbf{EDS}, we follow the evaluation setup of~\cite{burkhardt2025superevent} and report the Area Under the Curve (AUC) of pose estimation errors at ($5^\circ, 10^\circ, 20^\circ$), using rotation error. On \textbf{ECM} and \textbf{M3ED}, we follow the evaluation setup of~\cite{sun2021loftr}, where the pose error is defined as the maximum angular error in rotation and translation. To reduce randomness, we repeat the RANSAC five times and calculate the AUC over all errors. We also report matching precision following~\cite{sarlin2020superglue}, defined as the average ratio of the correct matches to total matches, where a correct match has an epipolar distance below $1\cdot 10^{-4}$ for outdoor scenes and $5\cdot 10^{-4}$ for indoor scenes. All images and events are resized so that the longer side is around 640 pixels. 
Unless otherwise specified, all models are evaluated using 128 ms inputs. Since the largest scale of SuperEvent input is 0.1 s, its input is also set to 128 ms. For MatchAnything and VGGT, we observed better performance with 30 ms input, which is therefore used as their evaluation interval. \\

\textbf{Baseline Methods.}
SuperEvent~\cite{burkhardt2025superevent} adopts a detector-based architecture that jointly predicts detections and descriptions, trained under supervision from image-based methods. 
MatchAnything~\cite{he2025matchanything} is trained on a large-scale self-labeled cross-modal dataset to generalize on unseen real-world cross-modality matching tasks. 
VGGT is a 1.2-billion-parameter model trained on a diverse dataset with 3D annotations, designed for geometric reasoning and uniform estimation of 3D properties within a scene. For methods originally trained on images, we report both zero–shot performance and the results using images reconstructed from events in order to provide a fair comparison. \\

\begin{figure}[t]
    \centering
    \includegraphics[trim=9cm 9.3cm 2.0cm 3.5cm,clip,width=\linewidth]{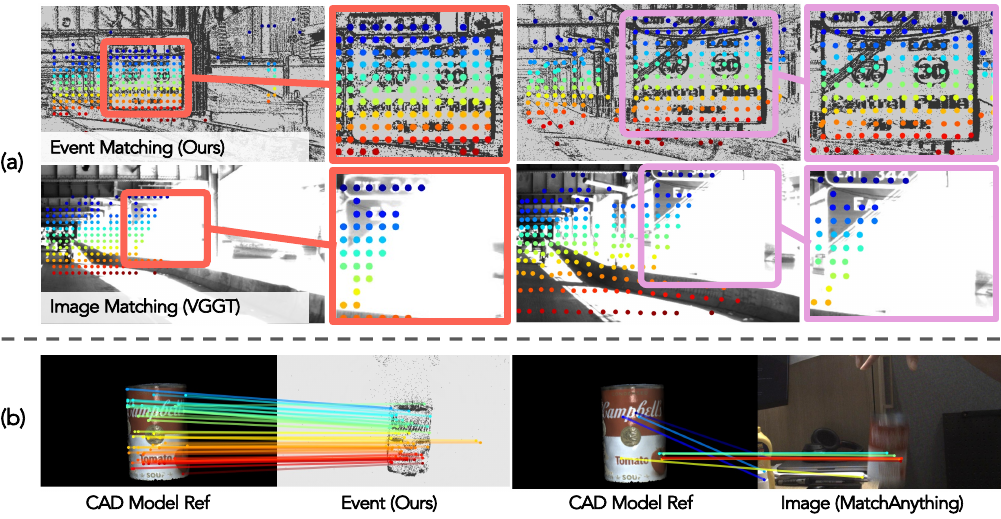}
    \caption{\textbf{Downstream tasks and applications}.
    \textbf{(a)} Event and image matching in HDR scene. Event cameras provide distinguishable visual details, enabling high-confidence matching where standard images fail due to extreme lighting. (The example shown is from \textit{schuylkill\_tunnel} in M3ED~\cite{chaney2023m3ed}) \textbf{(b)} Downstream applications for cross modality matching: Pose estimation for fast-moving objects.}
    \label{fig:downstream}
    \vspace{-.5cm}
\end{figure}
\begin{figure}[t]
    \centering
    \includegraphics[trim=6.7cm 10cm 6.5cm 7cm,clip,width=\linewidth]{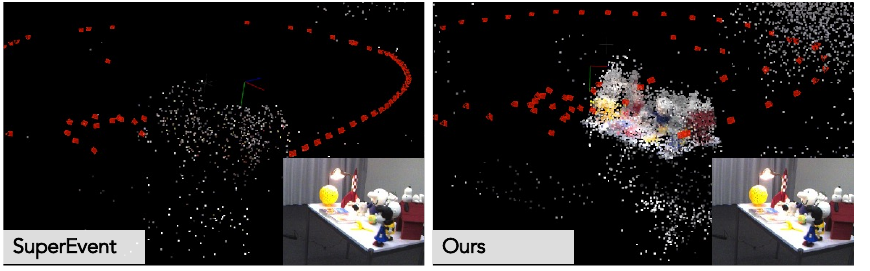}
    \caption{\textbf{Structure-from-Motion based on event matching.} \textbf{Left:} SuperEvent~\cite{burkhardt2025superevent}. \textbf{Right:} Ours.
    Our method produces more consistent camera poses and denser reconstruction. \textbf{Bottom Right of Each Image}: RGB frame for visualization reference.}
    \label{fig:sfm}
\end{figure}

\textbf{Implementation Details.}
Our model is trained on a single NVIDIA H100 GPU (96G) for three days with a batch size of 16 and an input resolution of $560\times336$. We use the AdamW optimizer with a learning rate of $6\cdot 10^{-4}$ and weight decay of $10^{-1}$. The ViT backbone is initialized with DINO\cite{simeoni2025dinov3} pretrained weights, as we empirically observed that this accelerates convergence. All remaining modules are trained from scratch. We set $N_l = 2$ and $N_r = 4$. The full model runs at 49ms for a $350\times630$ event pair on an NVIDIA RTX 4080 GPU.

\subsection{Results~\label{sec:e2e}}

\textbf{Results on ECM and M3ED.}
We evaluate the performance on our ECM and M3ED~\cite{chaney2023m3ed} test split, as is shown in~\cref{tab:main}. For ECM dataset, we build a sparse 3D reconstruction with incremental SfM pipeline in COLMAP~\cite{schoenberger2016sfm} and calculate overlap scores based on the number of common 3D points between views. We generate 462 pairs with minimum overlap score of 0.05 for evaluation. To fairly evaluate MatchAnything(M-A)~\cite{he2025matchanything} and VGGT~\cite{wang2025vggt} which are not trained on events, we provide event frames (Events) and intensity reconstruction from E2VID~\cite{rebecq2019high}. Since our model does not rely on long-range temporal dependencies, we disable the recurrent connections in E2VID for consistency. Results in~\cref{tab:main} show that our model learns robust matching predictions, offering \textbf{2.93 times} the AUC@5$^\circ$ than the state-of-the-art event-based baseline SuperEvent. We benchmarked computational efficiency on \ourdatashort{}. Our token selection mechanism reduces the FLOPs required for spatial attention operations by \textbf{21.5\%} (48.87G vs 62.22G) with minimal impact on overall performance. 
\\

\begin{table}[t]
\vspace{-0.3cm}
\centering
\caption{\label{tab:ablation} \textbf{Ablation study on ECM.}}
\begin{tabular}{cclcc}
\toprule
\textbf{Index} & \textbf{Ablation} & \textbf{Models} & \textbf{AUC@5°} & \textbf{Prec(\%)} \\
\midrule
\MakeUppercase{\romannumeral 1} & \multirow{3}{*}{Module} & w/o TAg & 50.12 & 67.18 \\
\MakeUppercase{\romannumeral 2} &                         & \MakeUppercase{\romannumeral 1} w/ Concentrate Net~\cite{nam2022stereo} & 48.37 & 64.79 \\
\MakeUppercase{\romannumeral 3} &                         & \MakeUppercase{\romannumeral 1} w/o Multi-scale Input & 39.86 & 66.87 \\
\midrule
\MakeUppercase{\romannumeral 4} & \multirow{2}{*}{Dataset} & Real Only(M3ED) & 11.33 & 20.25 \\
\MakeUppercase{\romannumeral 5} &                         & Syn Only(E-Mega) & 52.63 & 68.04 \\
\midrule
\MakeUppercase{\romannumeral 6} & -                       & \textbf{Full} & \textbf{54.61 }& \textbf{68.90} \\
\bottomrule
\end{tabular}
\vspace{-.5cm}
\end{table}

\textbf{Results on EDS~\cite{hidalgo2022event}.}
We evaluate our model on EDS using the test indices provided by~\cite{burkhardt2025superevent}. SuperEvent achieves higher performance after correcting the intrinsic parameters compared to the results reported in the original paper. Therefore, we report these improved values in \cref{tab:eds}. Due to the missing source code, we are unable to report results for EventPoint~\cite{huang2023eventpoint}. Our method achieves a \textbf{59\%} improvement over the detector-based SuperEvent. \\

\textbf{Downstream applications.} We show the effectiveness of our matching on the \textit{all\_characters} sequence in EDS dataset using \textbf{only events}. The event stream is temporally downsampled into 66 segments, each selected to ensure sufficient translation. We perform exhaustive pairwise matching across these event segments and reconstruct both camera poses and 3D point clouds using COLMAP, as shown in~\cref{fig:sfm}. Compared to SuperEvent, \ours{} produces significantly denser and clearer reconstructions and more consistent camera poses. Our matcher is capable of establishing reliable matches over large viewpoint changes, whereas SuperEvent struggles under large viewpoint changes and yields sparse keypoints. We also perform preliminary experiments
on two downstream applications in ~\cref{fig:downstream} (a) and (b). Event cameras allow us to match more robustly in challenging scenarios than
frame-based sensors due to dynamic range and motion blur.\\

\begin{figure}[t]
    \centering
    \begin{minipage}[b]{0.44\textwidth}
      \centering
        \includegraphics[trim=6.7cm 8.2cm 9.4cm 3.5cm,clip,width=1.0\linewidth]{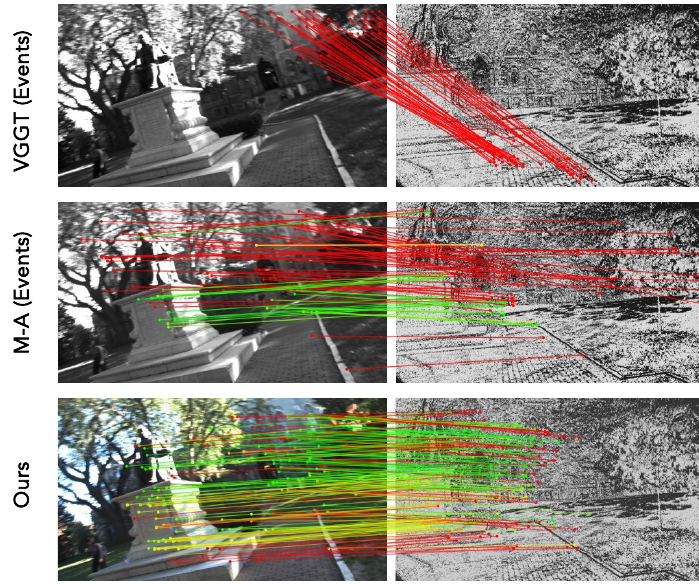}
        \caption{\textbf{Events-to-Image Matching}. \ours{} can match across wide baselines between images and events.}
        \label{fig:cross-matching}
    \end{minipage}
    \hspace{0.2cm}
    \begin{minipage}[b]{0.47\textwidth}
         \centering
        \includegraphics[trim=0.2cm 0.3cm 0.1cm 0.2cm,clip,width=\linewidth]{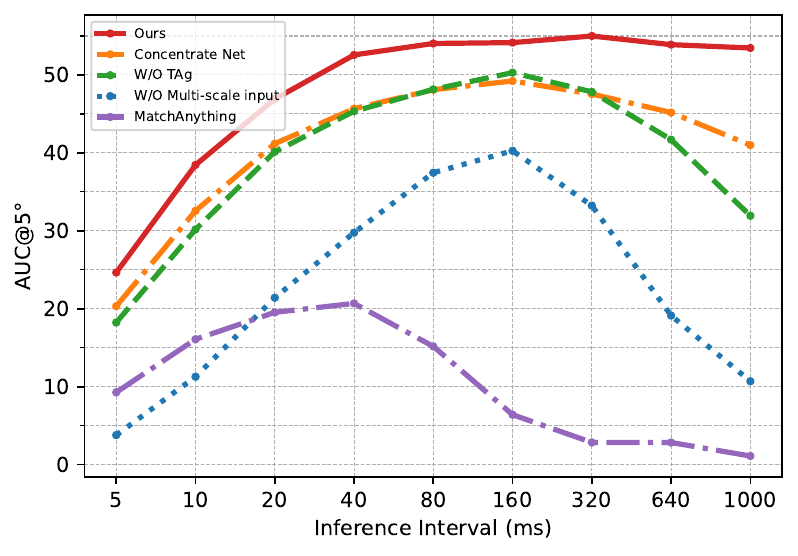}
        \caption{\textbf{Motion robustness study.} Four variants of Match-Any-Events and MatchAnything~\cite{he2025matchanything} are evaluated on ECM under a controlled interval setting, and the AUC@5$^\circ$ results are reported. 
        }
        \label{fig:motionrobust}
    \end{minipage}
    \vspace{-0.7cm}
\end{figure}

\textbf{Motion Robustness.}
We perform a thorough evaluation of motion robustness by measuring model performance while varying the input event window interval to simulate different motion speeds. Decreasing the interval below 20 ms results in a performance drop for all models, likely due to insufficient texture information. Quantitative results in~\cref{fig:motionrobust} demonstrate that our model maintains consistent performance across a wide range of interval values. Additionally, incorporating Concentrate Net~\cite{nam2022stereo} enhances robustness but it slightly reduces peak performance. In contrast, Match-Any-Events exhibits almost no performance degradation as the interval increases. \\
\label{sec:ablation}

\textbf{Ablation Study.} We evaluate four model variants in~\cref{tab:ablation}. Setup \MakeUppercase{\romannumeral 1} is trained without Temporal Aggregation Transformer (TAg). Instead, the temporal dimension is flattened into channels by multiplying the input dimension of the embedding layer by $N$.  Setup \MakeUppercase{\romannumeral 2} does not have multi-temporal scale input. Setup \MakeUppercase{\romannumeral 3} is built upon setup \MakeUppercase{\romannumeral 2} by replacing TAg with Concentrate Net~\cite{nam2022stereo}, a CNN that aggregates 3D volume into a 2D representation. We evaluated event- matching models trained on synthetic and real datasets in setup \MakeUppercase{\romannumeral 4} and \MakeUppercase{\romannumeral 5}. Models trained on M3ED show limited generalization, likely because driving scenes provide relatively constrained motion diversity. Setup \MakeUppercase{\romannumeral 6} is our full model. \\

\textbf{Event-to-Image Matching.} We further evaluate cross-modal matching on ECM and M3ED, both of which provide RGB images and events. As indicated in~\cref{tab:main}, MatchAnything~\cite{he2025matchanything} achieves decent performance in this task, while VGGT struggles to adapt to reconstructed images from events. and performs poorly on both datasets. In contrast, our method consistently outperforms all baselines, surpassing MatchAnything by about $7$ on AUC@5° on ECM and over $20$ on AUC@5° on M3ED. Qualitative comparisons in~\cref{fig:cross-matching} further demonstrate that our method produces more accurate correspondences across modalities.

\subsection{Analysis}
Our experiments show that the margin between our model and SuperEvent~\cite{burkhardt2025superevent} is larger on \ourdatashort{} than on M3ED~\cite{chaney2023m3ed}. M3ED offers pairs from adjacent views, whereas our benchmark samples views randomly, resulting in larger viewpoint changes and discontinuous motion. The gain comes from our multi-view synthetic training set, which provides diverse motions and long-range correspondence supervision. The Temporal Aggregation module further prevents overfitting to specific motion profiles and strengthens learning from synthetic data. We also find that VGGT (Events)~\cite{wang2025vggt} performs better on the indoor EDS dataset~\cite{hidalgo2022event}, likely because indoor scenes contain structured geometry, such as lines and planes, that benefits the three dimensional reasoning typical of foundation models.

%% file: sec/6_conclusion.tex
\section{Discussion}
We present \ours{}, a state-of-the-art generalizable matching model for event data. Our approach addresses two fundamental issues in event-based matching: 1) we introduce an efficient and scalable spatiotemporal transformer and a sparsity-aware event token selection module, and 2) we curate a large-scale synthetic dataset \ourdatasynth{} for wide-baseline matching and collect~\ourdata{}, a real-world challenging hetero-stereo dataset. By combining an efficient architecture and diverse data sources, we develop a generalizable matching model that handles wide baselines, motion variations, and cross-modality correspondences. \ours{} achieves state-of-the-art performance against all previous event matching methods, beating prior art by $37.7\%$ in zero-shot performance on unseen datasets.
\\
\\
\textbf{Acknowledgment.}
The financial support through the grants NSF FRR 2220868, NSF IIS 2212433, and ONR N00014-22-1-2677 is gratefully acknowledged. This work was also supported in part by funding from the Johns Hopkins Data Science and AI Institute.